\documentclass[conference]{IEEEtran}
\IEEEoverridecommandlockouts
\usepackage{cite}
\usepackage[breaklinks]{hyperref}
\hypersetup{colorlinks=true,
            linkcolor=blue,
            urlcolor=blue,
            anchorcolor=blue,
            citecolor=blue}
\usepackage{amsmath,amssymb,amsfonts}
\usepackage{algorithmic}
\usepackage{graphicx}
\usepackage{textcomp}
\usepackage{xcolor}
\usepackage{multirow}
\usepackage{adjustbox}

\def\BibTeX{{\rm B\kern-.05em{\sc i\kern-.025em b}\kern-.08em
    T\kern-.1667em\lower.7ex\hbox{E}\kern-.125emX}}
\begin{document}

\title{COCO-OLAC: A Benchmark for Occluded Panoptic Segmentation and Image Understanding
\thanks{$^*$: Equal contribution.}}

\author{\IEEEauthorblockN{Wenbo Wei$^*$}
\IEEEauthorblockA{\textit{Department of Computer Science} \\
\textit{University of Warwick, UK}\\
wenbo.wei@warwick.ac.uk}
\and
\IEEEauthorblockN{Jun Wang$^*$}
\IEEEauthorblockA{\textit{Department of Computer Science} \\
\textit{University of Warwick, UK}\\
jun.wang.3@warwick.ac.uk}
\and
\IEEEauthorblockN{Abhir Bhalerao}
\IEEEauthorblockA{\textit{Department of Computer Science} \\
\textit{University of Warwick, UK}\\
abhir.bhalerao@warwick.ac.uk}
}

\maketitle

\begin{abstract}
To help address the occlusion problem in panoptic segmentation and image understanding, this paper proposes a new large-scale dataset named COCO-OLAC (COCO Occlusion Labels for All Computer Vision Tasks), which is derived from the COCO dataset by manually labelling images into three perceived occlusion levels. Using COCO-OLAC, we systematically assess and quantify the impact of occlusion on panoptic segmentation for samples having different levels of occlusion. Comparative experiments with SOTA panoptic models demonstrate that the presence of occlusion significantly affects performance, with higher occlusion levels resulting in notably poorer performance. Additionally, we propose a straightforward yet effective method as an initial attempt to leverage the occlusion annotation using contrastive learning to develop a model that learns a more robust representation, capturing different severities of occlusion. Experimental results demonstrate that the proposed approach boosts the performance of the baseline model and achieves SOTA performance on the proposed COCO-OLAC dataset.\footnote{COCO-OLAC dataset and code are available at \href{https://github.com/wenbo-wei/COCO-OLAC}{https://github.com/wenbo-wei/COCO-OLAC}.}

\end{abstract}


\section{Introduction}
\begin{figure}
\centering
\setlength{\tabcolsep}{0.5pt}
\includegraphics[width=0.40\textwidth, height=0.26\textwidth]{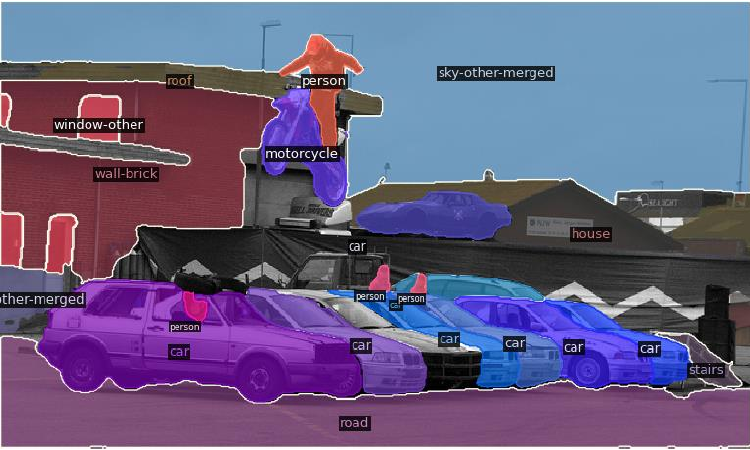}
\caption{Visualization of a prediction result by the Mask2Former \cite{cheng2022mask2former} network. The third car from the left in the middle, the truck in the center, and the truck near the roof are not detected due to occlusion.}
\label{fig:predict}
\vspace{-15pt}
\end{figure}

Owing to the release of large-scale datasets \cite{deng2009imagenet, zhou2017ade20k, cordts2016cityscapes, lin2014coco, neuhold2017mapillary, geiger2012kitti} and well-designed approaches, significant progress has been seen in computer vision \cite{szeliski2022cv} and image understanding tasks. Nonetheless, the performance of some tasks, e.g., object detection \cite{ren2016fasterrcnn, liu2016ssd, ross2017focalloss, lin2017featurepyramid, cai2018cascade, carion2020detr, zhu2020deformabledetr, liu2022dabdetr, zhang2022dino, wang2024yolov10}, instance segmentation \cite{he2017maskrcnn, liu2018path, bolya2019yolact, wang2020solov2, lee2020centermask}, and panoptic segmentation \cite{kirillov2019panoptic, kirillov2019panopticfpn, li2021panopticfcn, cheng2020panopticdeeplab, cheng2021maskformer, cheng2022mask2former, li2023maskdino, hu2023yoso, li2022pansegformer, jain2023oneformer}, still falls short of other tasks. The occlusion problem whereby some objects in the scene partially obscure others is one of the key and most common issues that hampers the further performance improvement on various computer vision tasks such as object detection and segmentation. This is because algorithms struggle to perceive the occluded object and adequately extract its features. Figure \ref{fig:predict} illustrates the occlusion problem for the panoptic segmentation task, where instances suffering from occlusion are wrongly predicted. To address this challenge, some studies have explored model architectures specially designed for handling occlusion in object detection~\cite{kortylewski2021compositional} and instance segmentation~\cite{yuan2021robustins, chen2015multiins, zhan2022trilayer, ke2021deepocclins}. For example, \cite{kortylewski2021compositional} introduces a Compositional Convolutional Neural Network (CompositionalNets), which enhances the robustness of deep convolutional neural networks for partial occlusion by integrating a differentiable compositional model. \cite{yuan2021robustins} generalises CompositionalNets to multi-object occlusion and proposes an Occlusion Reasoning Module (ORM) to post-hoc fix segmentation errors. 

Although some performance gains have been achieved, these methods normally are developed as intuitive or rule-based due to the lack of large-scale occlusion annotation data, which hinders potential further improvements. Therefore, in addition to the design of model architecture, datasets tailored for occlusion analysis are equally vital in research. \cite{qi2022ovis} proposes a benchmark dataset, OVIS, for video instance segmentation \cite{yang2019videoins} under severe occlusion. For each frame, they assign a unique occlusion level to the objects and define the Bounding-box Occlusion Rate (BOR) to qualify the occlusion level of the entire frame. \cite{ding2023mose} proposes MOSE, which partially inherits from OVIS, for video object segmentation in complex scenes. These datasets have promoted better methods for the video understanding task, but they are relatively small-scale and only cover a small part of commonly seen categories. Moreover, few studies have explored the occlusion in 2D image scenes, which is generally more challenging than 3D videos due to the absence of time series information \cite{kirillov2019panoptic}.

\begin{figure}
\centering
\setlength{\tabcolsep}{0.5pt}
\includegraphics[width=0.48\textwidth, height=0.31\textwidth]{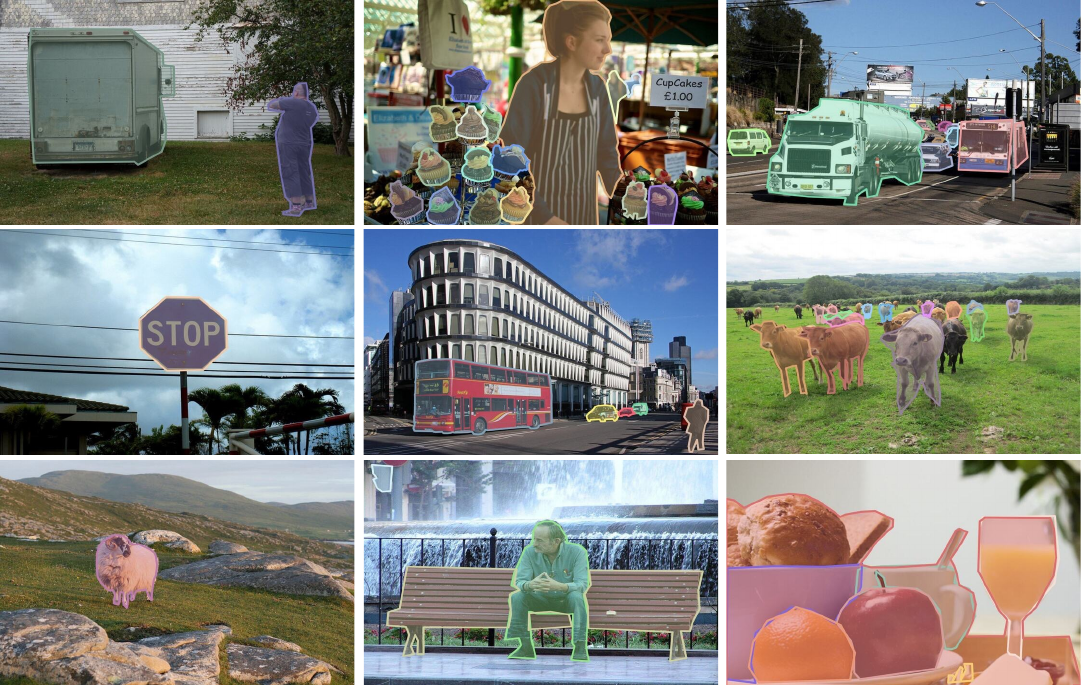}
\caption{Comparison of sample images with different occlusion levels from COCO-OLAC dataset. All images are overlaid with ground truth annotations to better show the occlusion relationship. The left, mid, and right columns show sample images with low, mid, and high occlusion levels, respectively. It is important to mention that only annotated instances are taken into account for assessment of the occlusion level.}
\label{fig:coco_occl_sample}
\vspace{-15pt}
\end{figure}

To fill this gap, we initiate our research by annotating the occlusion levels of images from the COCO dataset \cite{lin2014coco}. Current methods for occlusion measurement, such as BOR, define the occlusion level as the Intersection over Union (IoU) calculated by bounding boxes of scene instances. However, in most cases, a bounding box does not fully enclose the entire object, leading to BOR values that incorrectly represent the occlusion level. Moreover, BOR calculations heavily rely on the union area of the occluder and the occludee, neglecting the area in which the object is occluded, which further reduces its utility in occlusion measurement. To this end, although labour-intensive, we choose a direct approach and manually label the observed occlusion level (low, mid, and high), resulting in a new dataset named COCO-OLAC, as shown in Fig.~\ref{fig:coco_occl_sample}.  We then investigate the performance of recent SOTA methods \cite{kirillov2019panopticfpn, li2021panopticfcn, cheng2020panopticdeeplab, cheng2021maskformer, cheng2022mask2former, li2023maskdino} in understanding occluded images by exploring their performance in the challenging image understanding task, i.e., panoptic segmentation, on the proposed COCO-OLAC dataset. Note that although this work explores the proposed benchmark on a panoptic segmentation task, the proposed occlusion benchmark can be utilised to aid any image understanding tasks such as object detection and image recognition. Building on this, we further evaluate the newly trained Mask2Former \cite{cheng2022mask2former} across different occlusion levels. The experimental results show that performance decreases significantly as the occlusion level increases from low to high, underscoring the challenge that occlusion poses to SOTA panoptic segmentation methods.

As a first attempt to leverage the proposed occlusion annotations, we also propose a contrastive learning-based method to improve the representation learning of panoptic segmentation. Specifically, we utilise a triplet loss to bring samples having the same occlusion level closer in the feature space, while pushing apart samples with differing levels. Experiments show that the proposed method achieves state-of-the-art performance on the COCO-OLAC dataset. The ablation study further proves the effectiveness of the proposed method. 

In summary, our contributions are three-fold:
\begin{itemize}
\item We establish and release a large-scale dataset, COCO-OLAC, which aims to aid occluded image understanding tasks. This dataset is derived from the COCO dataset by manually annotating the observed occlusion level, resulting in 30,000 training images and 5,000 test images.
\item We systematically investigate the influence of the occlusion level by exploring the performance of SOTA panoptic segmentation methods in understanding the occluded images on the proposed benchmark.
\item We devise a method to leverage the proposed occlusion annotations using contrastive learning to improve the representation learning for the panoptic segmentation task. The experimental results demonstrate the effectiveness of the proposed method.
\end{itemize}

\section{Method}

\subsection{COCO-OLAC Dataset}\label{sec:coco_occl_dataset}

\paragraph{Dataset Annotation} 
 We annotate the occlusion levels of the images by the following three steps. For all the selected images, we first overlay polygon masks according to the ground truth annotation, since polygon masks can more accurately delineate the area of objects compared to bounding boxes. Then, we calculate the ratio of the occluded region to the entire occludee (including the occluded part), which is termed the occlusion rate. However, we cannot obtain the area of the occluded region directly because the COCO annotation only provides masks for visible parts of the objects. To resolve this issue, we have referred to a theory suggesting that people can estimate the shape of the occluded part of an object based on its overall shape, category, or relationship with other objects \cite{palmer1999visionscience}. Therefore, we opt to visually estimate the shape of the occluded region and thereby estimate the occlusion rate of the occludee. Finally, the highest occlusion rate is defined as the occlusion level of the image. Based on this criterion, we manually classify the images into three occlusion levels: low, mid, and high, corresponding to the occlusion rate of [0\%], (0\%, 50\%] and (50\%, 100\%], respectively. If the occlusion rate is difficult to estimate, which is often the case when determining the 50\% threshold, we classify such images as having a mid occlusion level.

\paragraph{Dataset Statistic} 
The COCO-OLAC dataset contains 35k images, with 30k training images selected from the original COCO train2017 (first 30k images) and 5k val2017 images, which are the same as the original COCO validation set. For the training set, 12,081 images are annotated as high occlusion, 11,251 images as mid occlusion, and 6,668 images as low occlusion. For the validation set, 1,791 images are annotated as high occlusion, 2,075 images as mid occlusion, and 1,134 images as low occlusion. In addition, we separate the validation set into three subsets according to the occlusion level for further experimental validation of the models.

\begin{figure}
\centering
\setlength{\tabcolsep}{0.5pt}
\includegraphics[width=0.45\textwidth]{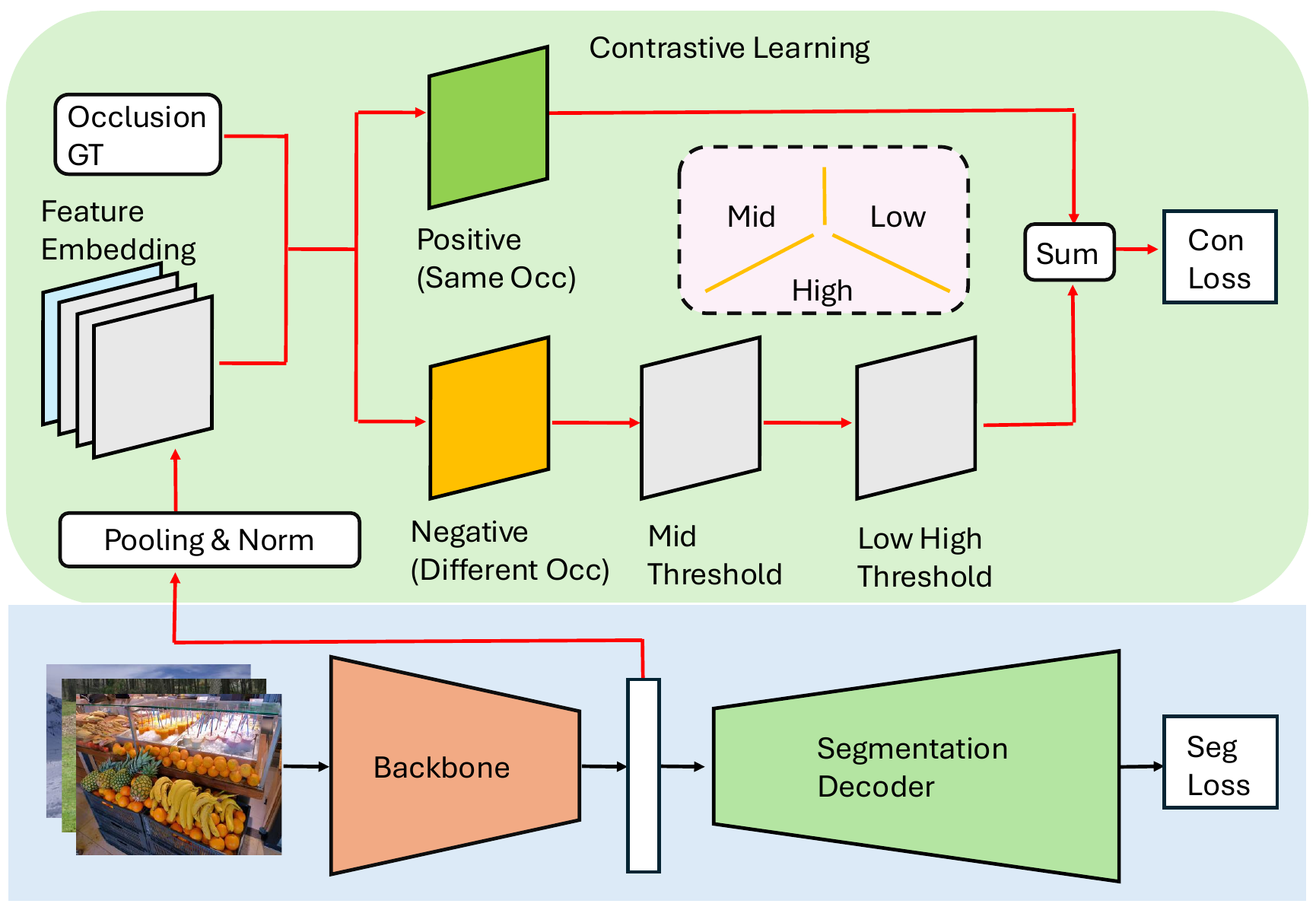}
\caption{The illustration of overall approach. The blue area below represents the schematic diagram of segmentation baseline, while the green area above illustrates our proposed contrastive learning-based method.}
\label{fig:contrastive}
\vspace{-15pt}
\end{figure}

\subsection{Contrastive Learning on Occlusion Level}\label{sec:con_learning}
Having the annotation of the occlusion level on hand, we aim to leverage this information to improve the model performance on occluded image understanding. The model is expected to recognize the different occlusion levels from different images, so as to learn a more robust feature representation. One simplest way is to form a classification task based on the occlusion level. However, the classification requires samples from the same occlusion level to be strictly mapped into the same latent space, ignoring the similarities among samples from different occlusion levels as the occlusion is a high-level concept. To this end, we employ a relatively soft method, the contrastive learning method, on the high-level image understanding task, i.e., panoptic segmentation, to improve the feature representation learning on occlusion. Specifically, we regard image pairs with the same occlusion level as positive pairs and those with different occlusion levels as negative pairs. A distance-based triplet loss is adopted to implement contrastive learning which could tolerate some dissimilarities between negative pairs.

The overall architecture is illustrated in Figure \ref{fig:contrastive}. In particular, we first feed the feature maps from the final layer of the backbone through $Norm(AvgPool(X))$ to obtain the feature embeddings $\tilde{\mathbf{z}}$. Then we calculate the cosine similarity by the function $Sim(\cdot)$ based on $\tilde{\mathbf{z}}$ between image samples within a batch and apply the triplet loss as shown in Equation \ref{eq:con_ls} to enable contrastive learning. Where $y_i$ denotes the occlusion label for the $i^{th}$ sample and $\tau$ is the margin.

\begin{align}
\label{eq:con_ls}
\begin{split}
 L_{con} = \frac{1}{B^2}\sum^{B}_{i=1}[\sum^{B}_{j:y^d_j=y^d_i}(1-\mathrm{Sim}(\tilde{\mathbf{z}_i},\tilde{\mathbf{z}_j})) \\
 +\sum^{B}_{j:y^d_j\not=y^d_i} \mathrm{max}(0,\mathrm{Sim}(\tilde{\mathbf{z}_i},\tilde{\mathbf{z}_j})-\tau)]
 \end{split}
\end{align}

We observe a noticeable difference between images with low and high occlusion levels. In contrast, images with mid occlusion levels share more similarities with those at low or high occlusion levels in the occlusion space. Given this, we propose to set two margins for the low-high image pairs and other pairs, respectively. Specifically, we set a strict threshold $\tau_{l,h}$ to push low-high image pairs closer and a higher threshold $\tau_{m}$ for other pairs to tolerate more dissimilarity. 

Through the proposed contrastive learning, the model can capture the occlusion difference from different samples, thus learning a more discriminative and robust feature representation. The model is jointly optimised by the proposed contrastive loss $L_{con}$ and the panoptic segmentation loss $L_{seg}$. The final loss $L_{fin}$ is shown in Equation \ref{eq:fin_loss} where $\lambda$ is a hyperparameter to balance the contribution between segmentation and contrastive learning.

\begin{equation}
\label{eq:fin_loss}
L_{fin} = L_{seg} + \lambda L_{con}
\end{equation}

\begin{table}
\centering
\caption{SOTA performance on different occlusion levels.}
\begin{adjustbox}{width=0.48\textwidth}
\begin{tabular}{c|c|c c c|c c}
\hline
\textbf{Method} & \textbf{Occlusion} & \textbf{PQ} & $\textbf{PQ}^{\textbf{Th}}$ & $\textbf{PQ}^{\textbf{St}}$ & $\textbf{AP}^{\text{Th}}_{\text{pan}}$ & $\textbf{mIoU}_{\text{pan}}$ \\ 
\hline
\multirow{3}{*}{Panoptic FPN \cite{kirillov2019panopticfpn}} & low & 43.8 & 53.2 & 29.5 & - & - \\
& mid & 40.2 & 47.3 & 29.5 & - & - \\
& high & 34.5 & 39.0 & 27.7 & - & - \\
\hline
\multirow{3}{*}{Panoptic FCN \cite{li2021panopticfcn}} & low & 46.9 & 56.1 & 33.3 & - & - \\
& mid & 41.9 & 48.2 & 32.5 & - & - \\
& high & 36.3 & 40.4 & 30.1 & - & - \\
\hline
\multirow{3}{*}{Panoptic DeepLab \cite{cheng2020panopticdeeplab}} & low & 42.9 & 47.8 & 35.5 & - & - \\
& mid & 36.2 & 39.4 & 31.3 & - & - \\
& high & 30.3 & 31.0 & 29.2 & - & - \\
\hline
\multirow{3}{*}{MaskFormer \cite{cheng2021maskformer}} & low & 52.6 & 58.8 & 43.3 & - & - \\
& mid & 48.0 & 53.9 & 39.1 & - & - \\
& high & 41.2 & 44.0 & 37.0 & - & - \\
\hline
\multirow{3}{*}{Mask2Former \cite{cheng2022mask2former}} & low & 56.8 & 64.4 & 45.8 & 56.5 & 60.4 \\
& mid & 53.3 & 60.1 & 43.0 & 45.1 & 61.2 \\
& high & 46.7 & 51.3 & 39.7 & 35.8 & 58.1 \\
\hline
\multirow{3}{*}{Mask DINO \cite{li2023maskdino}} & low & 56.6 & 63.1 & 47.0 & 56.4 & 58.0 \\
& mid & 53.7 & 60.6 & 43.4 & 47.2 & 59.7 \\
& high & 48.3 & 53.3 & 40.8 & 38.8 & 57.4 \\
\hline
\end{tabular}
\label{table:eval_across}
\end{adjustbox}
\end{table}

\begin{table}
\caption{Comparison of SOTA Panoptic Segmentation Methods on the COCO-OLAC Dataset via retraining on COCO-OLAC dataset.}
\begin{adjustbox}{width=0.48\textwidth}
\begin{tabular}{c|c c c|c c}
\hline
\textbf{Model} & \textbf{PQ} & $\textbf{PQ}^{\textbf{Th}}$ & $\textbf{PQ}^{\textbf{St}}$ & $\textbf{AP}^{\text{Th}}_{\text{pan}}$ & $\textbf{mIoU}_{\text{pan}}$ \\ 
\hline
Panoptic FPN \cite{kirillov2019panopticfpn} & 33.3 & 39.1 & 24.6 & - & - \\
Panoptic-DeepLab \cite{cheng2020panopticdeeplab} & 27.5 & 28.9 & 25.3 & - & -\\
Panoptic FCN \cite{li2021panopticfcn} & 33.0 & 37.7 & 26.0 & - & - \\
Mask2Former \cite{cheng2022mask2former} & 41.5 & 45.3 & 35.6 & 30.6 & \textbf{54.4} \\
YOSO \cite{hu2023yoso} & 37.1 & 40.6 & 31.9 & - & - \\
\hline
Ours & \textbf{41.8} & \textbf{45.3} & \textbf{36.4} & \textbf{30.8} & 54.3 \\
\hline
\end{tabular}
\label{table:comp_sota_methods}
\end{adjustbox}
\end{table}

\section{Experiments}
We investigate how images with different occlusion levels affect the performance of the SOTA panoptic segmentation method using the newly proposed COCO-OLAC dataset. Apart from the standard panoptic segmentation metric PQ \cite{kirillov2019panoptic}, we also report $\text{AP}^{\text{Th}}_{\text{pan}}$ and $\text{mIoU}_{\text{pan}}$ \cite{cheng2022mask2former} for instance segmentation and semantic segmentation, respectively. Finally, we demonstrate the effectiveness of our proposed methods through ablation studies.

\subsection{Implementation Details}\label{sec:implementation_details}
For experiments investigating the influence of different occlusion levels, we conduct a validation experiment that loads the pre-trained weights (trained on the entire COCO dataset) of recent SOTA panoptic segmentation methods and tests their performance on the different validation subsets (occlusion level) of the proposed COCO-OLAC using the official code and scripts. For experiments verifying the effectiveness of our proposed method, we adopt Mask2Former (with ResNet-50) as our baseline and train the model on the training set of our proposed COCO-OLAC dataset. All the compared methods on this experiment are reimplemented on the COCO-OLAC dataset using the official code. The input images are resized and cropped into $512 \times 512$. $\lambda$ in equation \ref{eq:fin_loss} is set to 1.0. The margins $\tau_{l,h}$ and $\tau_m$ are set to 0.4 and 0.6 respectively.

\subsection{Experimental Results}
Herein, we first benchmark the performance of the SOTA panoptic segmentation methods, including Panoptic FPN \cite{kirillov2019panopticfpn}, Panoptic DeepLab \cite{cheng2020panopticdeeplab}, Panoptic FCN \cite{li2021panopticfcn}, Mask2Former \cite{cheng2022mask2former} and Mask DINO \cite{li2023maskdino}, on the proposed COCO-OLAC validation set. In particular, we report the results on different levels of occlusion (subsets), as shown in Table \ref{table:eval_across}. The results show a clear trend: As the level of occlusion increases from low to high, the performance metrics (PQ, $\text{PQ}^{\text{Th}}$, $\text{PQ}^{\text{St}}$, and $\text{AP}^{\text{Th}}_{\text{pan}}$) experience a significant decline. This degradation highlights a critical limitation of current SOTA panoptic segmentation methods, which struggle to maintain accuracy and consistency in handling occlusion. Moreover, it also confirms the correctness of our manual annotation that reflects the level of occlusion of the image.  

Then, we verify the effectiveness of our proposed method by retraining and testing the model on the COCO-OLAC. We demonstrate the results of our proposed method and the baseline on all three levels of occlusion subsets of the validation set in Table \ref{table:comp_across} and compare our method with recent SOTA methods in Table \ref{table:comp_sota_methods}. As can be seen, the model equipped with the proposed contrastive learning shows substantial improvements at all levels of occlusion. Specifically, for low occlusion, PQ improves by 0.6, and $\text{PQ}^{\text{St}}$ sees a notable increase of 2.3, indicating a strong improvement in the segmentation of 'stuff' classes. Similarly, under high occlusion, the method produces a 0.4 improvement in PQ and a 1.0 increase in $\text{PQ}^{\text{St}}$. All results demonstrate the effectiveness of our method compared to the baseline in a range of occlusion levels. Note that the panoptic segmentation is a challenging task, especially on the large-scale challenging COCO dataset, where even a small improvement proves difficult. Moreover, our method obtains the best overall performance over previous SOTA methods. These results confirm our theory and demonstrate that the representation of the occlusion feature can assist in training and improve overall segmentation accuracy, especially in segmenting and labelling larger and less defined regions.

\begin{table}
\centering
\caption{Comparison Across Low, Mid, and High Occlusion Levels for Proposed Methods and Baseline via retraining.}
\begin{adjustbox}{width=0.48\textwidth}
\begin{tabular}{c|c|c c c|c c}
\hline
\textbf{Occlusion} & \textbf{Model} & \textbf{PQ} & $\textbf{PQ}^{\textbf{Th}}$ & $\textbf{PQ}^{\textbf{St}}$ & $\textbf{AP}^{\text{Th}}_{\text{pan}}$ & $\textbf{mIoU}_{\text{pan}}$ \\ 
\hline
\multirow{2}{*}{low} & base & 47.5 & \textbf{53.5} & 38.5 & \textbf{46.7} & 52.2 \\
& con & \textbf{48.1}(+0.6) & 53.1 & \textbf{40.8}(+2.3) & 46.4 & \textbf{52.7}(+0.5) \\
\hline
\multirow{2}{*}{mid} & base & 43.1 & \textbf{48.1} & 35.6 & 33.3 & \textbf{54.0} \\
& con & \textbf{43.2} & 47.9 & \textbf{36.1}(+0.5) & \textbf{33.7}(+0.4) & 54.0 \\
\hline
\multirow{2}{*}{high} & base & 35.7 & 38.2 & 32.0 & \textbf{24.8} & 50.7 \\
& con & \textbf{36.1}(+0.4) & \textbf{38.2} & \textbf{33.0}(+1.0) & 24.7 & \textbf{50.8} \\
\hline
\end{tabular}
\label{table:comp_across}
\end{adjustbox}
\vspace{-10pt}
\end{table}

\begin{table}
\caption{Ablation study on the proposed method.}
\begin{center}
\begin{tabular}{c|c c c|c c}
\hline
\textbf{Model} & \textbf{PQ} & $\textbf{PQ}^{\textbf{Th}}$ & $\textbf{PQ}^{\textbf{St}}$ & $\textbf{AP}^{\text{Th}}_{\text{pan}}$ & $\textbf{mIoU}_{\text{pan}}$ \\ 
\hline
Base & 41.5 & 45.3 & 35.6 & 30.6 & \textbf{54.4} \\
Our & \textbf{41.8}(+0.3) & \textbf{45.3} & \textbf{36.4}(+0.8) & \textbf{30.8} & 54.3 \\
\hline
\end{tabular}
\label{table:con_ablation}
\end{center}
\end{table}

\subsection{Ablation Study}
In addition to the ablation study in Table \ref{table:comp_across}, we further conduct an ablation study to analyse the effectiveness of our method on the entire validation set.

As shown in \ref{table:con_ablation}, after adding the proposed method, an improvement can be seen on both the various occlusion subsets and the entire validation set, especially $\text{PQ}^{\text{St}}$, which improves by 0.8 on the entire validation set. This indicates that the proposed contrastive learning method enhances the model's ability to learn a more occlusion-robust feature, thereby improving its overall performance.

Then, we investigate the influence of the margins in our proposed contrastive learning method in Table \ref{table:threshold_comb}. In particular, we first evaluate the combination of thresholds of $\tau_{l,h}$ and $\tau_{m}$ by keeping one threshold fixed while increasing the other. When the first threshold $\tau_{l,h}$ is fixed at 0.3, increasing the second threshold $\tau_m$ (from 0.4 to 0.8) shows a minimal effect on overall performance. When $\tau_{l,h}$ increases to 0.4, the combination of $\tau_{l,h} = 0.4$ and $\tau_m = 0.6$ achieves the highest PQ. However, with further increases of $\tau_m$ beyond 0.6, the overall performance starts to decline. This trend demonstrates that neglecting the influence of either threshold will result in a degradation of overall performance.

Our method is the initial attempt to utilise the occlusion annotations. We believe that a more significant improvement can be achieved when further advanced methods are proposed to better optimise our proposed dataset.

\begin{table}
\caption{Impact of $\tau_{l,h}$ and $\tau_{m}$ Combinations on the Performance of Contrastive Learning-Based Method}
\begin{center}
\begin{tabular}{c c|c c c}
\hline
$\tau_{l,h}$ & $\tau_{m}$ & $\textbf{PQ}$ & $\textbf{PQ}^{\textbf{Th}}$ & $\textbf{PQ}^{\textbf{St}}$ \\ 
\hline
0.3 & 0.4 & 41.3 & 45.0 & 35.8 \\
0.3 & 0.6 & 41.2 & 44.9 & 35.6 \\
0.3 & 0.8 & 41.4 & 45.2 & 35.7 \\
0.4 & 0.6 & \textbf{41.8} & \textbf{45.3} & \textbf{36.4} \\
0.4 & 0.7 & 41.3 & 45.1 & 35.7 \\
\hline
\end{tabular}
\label{table:threshold_comb}
\end{center}
\vspace{-10pt}
\end{table}

\section{Conclusion}
In this work, we introduce COCO-OLAC, a novel large-scale dataset designed to facilitate research on occlusion in panoptic segmentation and other image understanding tasks. The dataset, derived from the COCO dataset with manual occlusion annotations, provides a comprehensive benchmark with over 30,000 training images and 5,000 test images categorised into three occlusion levels. Our experiments systematically evaluated the performance of SOTA panoptic segmentation methods on this dataset, revealing that occlusion, especially at higher levels, substantially degrades model performance. In addition, we propose a contrastive learning-based method to enhance model robustness in handling occlusion. Despite its simplicity, our approach yields performance improvements over the baseline. This work not only highlights the significant impact of occlusion on the panoptic segmentation task but also offers a quantitative foundation for future research into more effective occlusion-aware methods.

\bibliographystyle{IEEEtran}
\bibliography{references}

\end{document}